\documentclass[10pt,twocolumn,letterpaper]{article}

\usepackage{cvpr}
\usepackage{times}
\usepackage{epsfig}
\usepackage{graphicx}
\usepackage{subcaption}
\usepackage{amsmath}
\usepackage{amssymb}
\usepackage{multirow}
\usepackage{enumitem}
\usepackage{bbm}

\usepackage[breaklinks=true,bookmarks=false]{hyperref}

\cvprfinalcopy 

\def\cvprPaperID{1747} 

\ifcvprfinal\fi
\begin{document}

\title{Walk and Learn: Facial Attribute Representation Learning from Egocentric Video and Contextual Data}

\author{Jing Wang\\
Northwestern University\\
{\tt\small jing.wang@u.northwestern.edu}
\and
Yu Cheng\\
IBM T. J. Watson\\
{\tt\small chengyu@us.ibm.com}
\and 
Rogerio Schmidt Feris\\
IBM T. J. Watson\\
{\tt\small rsferis@us.ibm.com}
}

\maketitle
\thispagestyle{empty}

\begin{abstract}The way people look in terms of facial attributes (ethnicity, hair color, facial hair, etc.) and the clothes or accessories they wear (sunglasses, hat, hoodies, etc.) is highly dependent on geo-location and weather condition, respectively. This work explores, for the first time, the use of this contextual information, as people with wearable cameras walk across different neighborhoods of a city, in order to learn a rich feature representation for facial attribute classification, without the costly manual annotation required by previous methods. By tracking the faces of casual walkers on more than 40 hours of egocentric video, we are able to cover tens of thousands of different identities and automatically extract nearly 5 million pairs of images connected by or from different face tracks, along with their weather and location context, under pose and lighting variations. These image pairs are then fed into a deep network that preserves similarity of images connected by the same track, in order to capture identity-related attribute features, and optimizes for location and weather prediction to capture additional facial attribute features. Finally, the network is fine-tuned with manually annotated samples. We perform an extensive experimental analysis on wearable data and two standard benchmark datasets based on web images (LFWA and CelebA). Our method outperforms by a large margin a network trained from scratch. Moreover, even without using manually annotated identity labels for pre-training as in previous methods, our approach achieves results that are better than the state of the art.
\end{abstract}

\section{Introduction}

Describing people based on attributes, such as gender, age, hair style and clothing style, is an important problem for many applications, including suspect search based on eyewitness descriptions \cite{Feris:2014:APS:2578726.2578732}, fashion analytics \cite{liu2012street,kiapourbuy,chen2015deep}, face retrieval and verification \cite{twobirdsonestone,berg-poof-cvpr2013,liu2015faceattributes}, and person re-identification \cite{layne2014re,DBLP:conf/cvpr/ShiHX15}. In this work, we address the problem of learning rich visual representations (i.e., ``good features") for modeling person attributes without manual labels, with a focus on facial attribute prediction.

The state of the art in facial attribute classification, as demonstrated by standard evaluation benchmarks \cite{liu2015faceattributes}, has been advanced by methods that use deep convolutional neural networks (CNNs) pre-trained on massive amounts of images that have been manually annotated with identity labels. In fact, it has been shown that identity-related attributes such as gender, hair color, and age are implicitly encoded in nodes of CNNs that are trained for identity discrimination \cite{liu2015faceattributes,DBLP:conf/cvpr/SunWT15}. Despite the excellent performance, the feature representation learned by these methods requires {\em costly manual annotation} of hundreds of thousands or even millions of images in the pre-training stage.  Moreover, the pre-trained network fails to encode attributes that are not related to identity, such as eyewear and different types of hats.

\begin{figure*}
\centering
\includegraphics[width=0.9\textwidth]{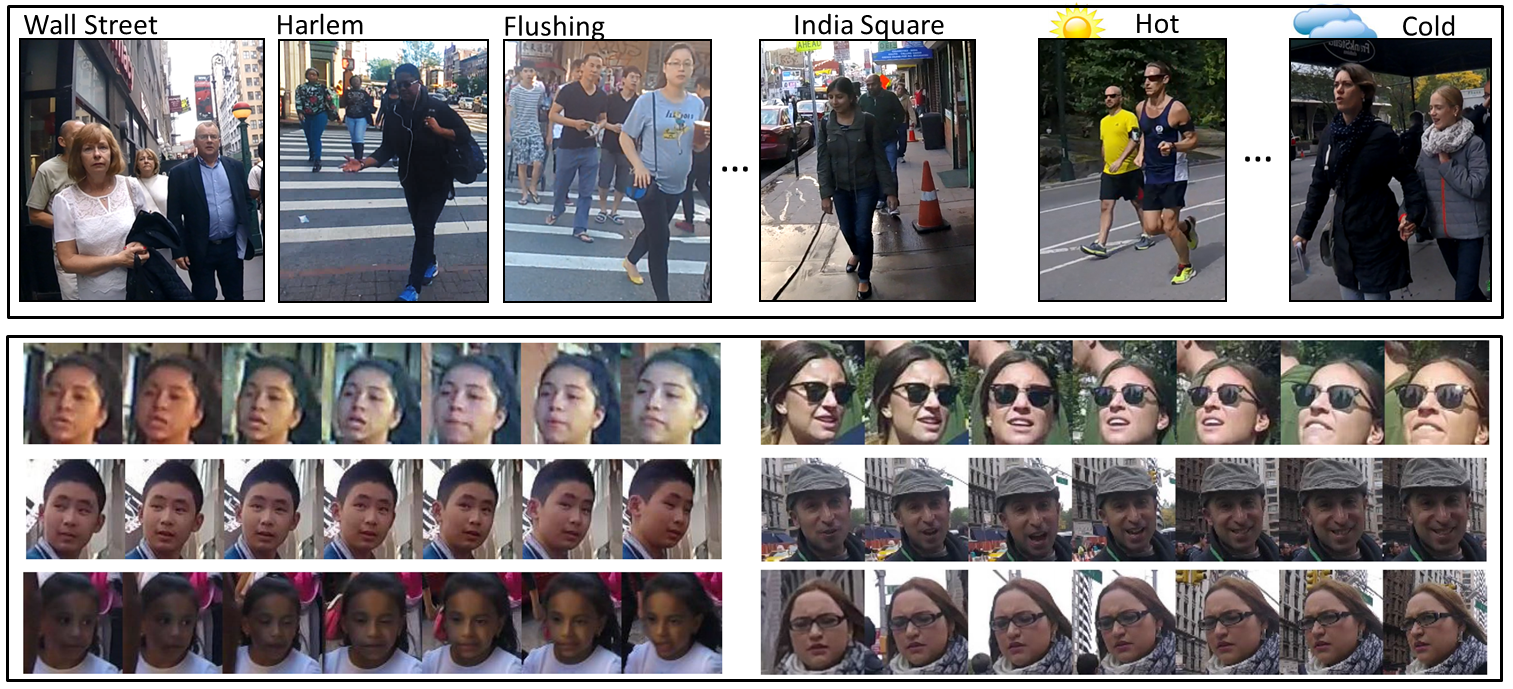}
\caption{{\bf Top:} Casual walkers as imaged by people with wearable cameras walking across different neighborhoods of New York City. Due to changes in demographics, the expected appearance of facial attributes is highly dependent on location. Moreover, the weather conditions change facial appearance due to different lighting, and influence the choice of outfit and the use of accessories such as sunglasses, hats, and scarfs. {\bf Bottom:} Face images obtained via face detection and landmark tracking. Note the large variations in lighting, expression, face pose, ethnicities, and accessories. We exploit this information to build rich visual feature representations for facial attribute classification. }
\label{fig:fig1}
\end{figure*}

In this paper, we address these issues by taking a different approach. Instead of relying on manually annotated images from the web, we learn a discriminative facial attribute representation from {\em egocentric videos} captured by a person walking across different neighborhoods of a city, while leveraging discretized geo-location and weather information readily available in wearable devices as a free source of supervision.  
The motivation for using location and weather data to construct facial attribute representations is illustrated in Figure \ref{fig:fig1}. In New York City, for example, the likelihood of meeting an Afro-American casual walker in certain regions of Harlem is more than 90\%. The same is true for Hispanics in Washington Heights, East Asians in Flushing, South Asians in India Square, East Europeans in Brighton Beach, and so on. These groups are characterized by their unique facial attributes (hair color, hair length, facial and eyes shape, etc.). Moreover, the weather conditions influence the facial appearance changes due to lighting variations and also dictate the clothing and accessories people wear. As an example, on sunny and warm days, the likelihood that a person will wear sunglasses, baseball hats, t-shirts, and shorts increases, whereas the presence of scarfs, beanies, and jackets is much more frequent in cold days. Our goal is to leverage data about location and weather as weak labels to construct rich facial attribute representations.

\vspace{0.02in}
{\bf Overview of our Approach.} Our proposed feature learning method relies on processing identity-unlabeled data and learning feature embeddings from a few supervised tasks.
We first track the faces of casual walkers using facial landmark tracking in more than 40 hours of egocentric video, obtaining face images under a variety of conditions, as shown in Figure \ref{fig:fig1}. These face images are then arranged into pairs, where information from tracking is used to label the pairs as belonging to the same individual or not. Nearly 5 million pairs are generated and fed into a network that encodes identity-related features through a Siamese structure with contrastive loss, while further embedding contextual features based on location and weather prediction. Finally, the obtained feature representation is fine-tuned with manual labels for the task of facial attribute classification.

Generally, our proposed feature representation learning for person attribute modeling has the following advantages over previous methods: First, it does not require costly manual annotation in the pre-training stage. Second, by leveraging location and weather information, it encodes facial features beyond identity, in contrast to methods pre-trained on large image repositories with identity labels \cite{berg-poof-cvpr2013,liu2015faceattributes}. Third, it leverages the rich appearance of faces from a large number of casual walkers at different locations and lighting conditions, which may not be captured by images available on the web.

\vspace{0.02in}
Our main {\bf contributions} can be summarized as follows: 

\vspace{0.02in}
1. We introduce a new {\em Ego-Humans dataset} containing more than 40 hours of egocentric videos captured by people with wearable cameras walking across different regions of New York City. The data covers tens of thousands of casual walkers and includes both the weather and location context associated with the videos. 

\vspace{0.02in}
2. To the best of our knowledge, this is the first time a ``walk and learn" approach that leverages discretized geo-location and weather information has been proposed for constructing deep visual representations for person attribute modeling. Our method seamlessly embeds this contextual information in a Siamese network that measures similarity of face pairs automatically extracted from tracks. 

\vspace{0.02in}
3. We show that our {\em self-supervised} approach can match or exceed the performance of state-of-the-art methods that rely on supervised pre-training based on hundreds of thousands or millions of annotated images with identity labels. In addition, we show that facial attributes are implicitly encoded in our network nodes as we optimize for location, weather, and face similarity prediction.

\section{Related Work}
{\bf Egocentric Vision.} First-person vision methods have received renewed attention by the computer vision community \cite{DBLP:journals/ijcv/LeeG15,Li_2015_CVPR,Rogez_2015_CVPR}. Current methods and datasets have focused on problems such as video summarization \cite{DBLP:journals/ijcv/LeeG15,rita2015}, activity recognition \cite{Fathi:2012:LRD:2402940.2402964}, and social interaction \cite{DBLP:conf/cvpr/FathiHR12}. In contrast, our work is focused on the problem of {\em looking at people} and modeling facial attributes from a first-person vision perspective. Compared to existing egocentric datasets \cite{DBLP:journals/tcsv/BetancourtMRR15}, our Ego-Humans dataset is the first of its kind; it deals with a different task, it is larger in scale, and it also has associated geo-location and weather information, which could be relevant for many other tasks.

{\bf Facial Attribute Modeling.}
Kumar et al. \cite{kumar2011describable} proposed a method based on describable facial attributes to assist in face verification and attribute-based face search. Siddiquie et al. \cite{Siddiquie:2011:IRR:2191740.2192123} and Luo et al. \cite{Luo:2013:DSA:2586117.2587104} exploited the inter-dependencies of facial attributes to improve classification accuracy. Chen et al. \cite{hchen_gallagher_girod_cvpr_13_firstnames} built a feature representation that relies on discrimination of images based on first names, and showed improved results in age and gender classification. Berg and Belhumeur \cite{berg-poof-cvpr2013} introduced part-based one-vs.-one features (POOFs) and showed that features constructed based on identity discrimination are helpful for facial feature classification. Li et al. \cite{twobirdsonestone} proposed a method that jointly learns discriminative binary codes and attribute prediction for face retrieval.

More recently, deep convolutional neural networks have advanced the state of the art in facial attribute classification. N. Zhang et al. \cite{zhang2014panda} proposed pose-aligned networks (PANDA) for deep attribute modeling. Z. Zhang et al. \cite{socialrelation_ICCV2015} proposed a deep model based on facial attributes to perform pairwise face reasoning for social relation prediction. Lin et al. \cite{liu2015faceattributes} achieved state-of-the-art performance on the LFWA and CelebA datasets using a network pre-trained on massive identity labels. Our work, instead, achieves the same or superior performance without requiring manually annotated identity labels for the pre-training step.

{\bf Geo-Tagged Image Analysis.}
Many methods have been proposed for geo-tagged image analysis. In particular, image geo-localization, i.e., the problem of predicting the location of a query image, has received increased attention in the past few years \cite{hays2008im2gps,gronat2013learning,lin2013cross,lee2015predicting}. Other related research includes discovering architectural elements and recognizing city attributes from large geo-tagged image repositories \cite{Doersch:2012:MPL:2185520.2185597,LNCS86910519} and using location context to improve image classification \cite{TangPFFB15}. More closely related to our work,  Islam et al. \cite{geofacial,islam2015face2gps} investigated the geo-dependence of facial features and attributes; however they used off-the-shelf facial attribute classifiers for this analysis, whereas the goal of our work is to build feature representations so as to improve the accuracy of facial attribute classifiers.  

{\bf Representation Learning.}
Most high-performance computer vision methods based on deep learning rely on visual representations that are learned based on {\em supervised pre-training}, for example, using networks trained on millions of annotated examples such as the ImageNet dataset for general object classification \cite{imagenet_cvpr09,krizhevsky2012imagenet}, or relying on massive amounts of identity labels for facial analysis tasks \cite{liu2015faceattributes,DBLP:conf/cvpr/SunWT15}. Our work, instead, is focused on building rich visual representations for person attribute classification without using manual annotations in the pre-training step.

There is a long history of methods for unsupervised learning of visual representations based on deep learning \cite{Hinton:2006:FLA:1161603.1161605,Zeiler10deconvolutionalnetworks,DBLP:journals/corr/ZhaoMGL15}.  When large collections of unlabeled still images are available, auto-encoders or methods that optimize for reconstruction of the data are popular solutions to learn features without manual labeling \cite{DBLP:conf/icml/LeRMDCCDN12,DBLP:journals/corr/ZhaoMGL15,Vincent:2008:ECR:1390156.1390294}. 
Doersch et al. \cite{doersch2015unsupervised} proposed learning supervised ``pretext'' tasks between patches within an image as an embedding for unsupervised object discovery.
These approaches, however, have not yet proven effective in matching the performance of supervised pre-training methods.

When video data is available, additional regularization can be imposed by enforcing temporal coherence \cite{Mobahi:2009:DLT:1553374.1553469,DBLP:journals/corr/WangG15a} or through the so called slow feature analysis \cite{Wiskott:2002:SFA:638940.638941}. More recently, Srivastava et al. \cite{DBLP:conf/icml/SrivastavaMS15} used multilayer Long Short Term Memory (LSTM) networks to learn representations of video sequences, combining auto-encoders and prediction of future video frames.

Our work is related to other methods that learn visual representations from videos captured by wearable cameras or vehicle-mounted cameras \cite{DBLP:journals/corr/JayaramanG15,LSM2015}, where awareness of egomotion can be used as a supervisory signal for feature learning. In contrast to those methods, however, we leverage the geo-location and weather data that are readily available in wearable sensors as a source of free supervisory signal to learn rich visual representations which are suitable to facial attribute classification.


\section{Ego-Humans Dataset}
\label{sec:dataset}
\textbf{Data Collection.} The Ego-Humans dataset was collected in New York City over a period of two months, from August 28 to October 26 (during the summer and fall seasons). The data consists of videos captured by three people with chest-mounted cameras, walking across different neighborhoods of the city. Two camera models were used: a GoPro camera (higher-quality) and a Vievu camera (lower-quality), both with 1080p resolution, capturing data at 30 frames per second.
Within the two-month period, 25 days were selected for data collection, covering different regions of Manhattan and nearby areas, including the Financial District, Times Square, Central Park, Harlem, Little Italy, Brooklyn Bridge, Chinatown, Flushing, and others. In each day, one or more hours of video were recorded, at different times of the day and in different weather conditions. In total, we recorded more than 40 hours of egocentric videos, split into chunks of 15 minutes. In association with these videos, we recorded location using a GPS sensor and detailed weather information, such as temperature, precipitation, and weather condition, using an open weather API that retrieves this information based on geographic coordinates.

\begin{figure}
\centering 
\includegraphics[width=0.4\textwidth]{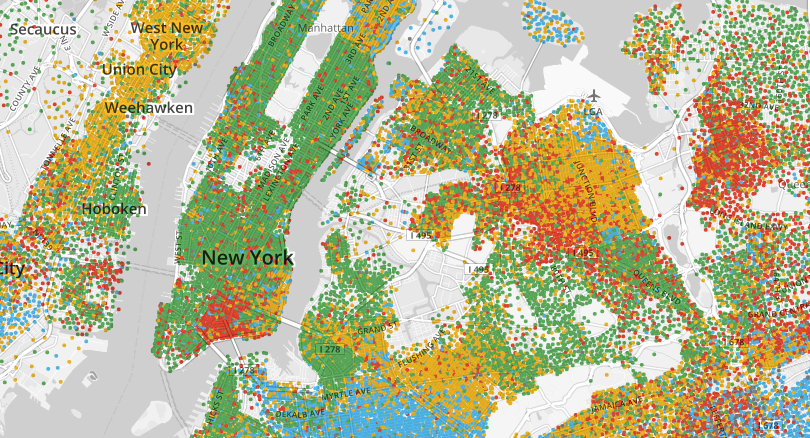}
\caption{Grouping of GPS coordinates based on an ethnicity map defined by census data (best viewed in color).}
\label{fig:map}
\end{figure}

\textbf{Discretization of Contextual Data.} 
Rather than relying on fine-grained GPS coordinates, our learning algorithm considers a coarse set of locations as class labels. More specifically, we cluster GPS coordinates according to published census/ethnicity data \footnote{http://projects.nytimes.com/census/2010/explorer}. In particular, we consider four ethnical groups: {\em White, Black, Asian, and Indian}. Figure \ref{fig:map} shows an ethnicity map segmented based on census data, where each cluster has its own peculiar predominance of facial attributes.  We are currently expanding this set (including Hispanics, for example) as we capture more data in other locations. Regarding weather, our data includes a variety of temperatures and conditions, but for training we have used two classes: {\em sunny/hot and cloudy/cold}. We note that other partitions of our data could be used for other tasks. As an example, for clothing attributes, GPS clustering based on socio-economic factors could be relevant, as well as finer-grained weather conditions and temperatures.

In addition to extracting the weather and location labels, it is also important to generate face pairs (similar and dissimilar) for encoding identity features, which are helpful for discriminating several facial attributes. This procedure consists of two steps: 1) tracking casual walkers via face detection and landmark tracking and 2) image pair selection. 

\textbf{Tracking Casual Walkers.}
We used the OpenCV frontal face detector and facial landmark tracking based on the supervised descent method (SDM) \cite{xiong2013supervised} to track casual walkers in the videos. The detector was tuned to output only high-confidence detections, with virtually no false alarms, at the expense of more false negatives. We used the {\em intraface} implementation of the  SDM landmark tracking \footnote{ http://www.humansensing.cs.cmu.edu/intraface/}, which  works remarkably well, greatly expanding the set of captured face poses, lighting, and expressions as illustrated in Figure \ref{fig:fig1}, without drifting. In total, we collected 15,000 face tracks, for a total of 160,000 face images.

\textbf{Selecting Informative Pairwise Constrains}: Given the face images extracted by face detection and tracking, we consider the following pairwise constraints:
\begin{itemize}[noitemsep,topsep=0pt]
\item Temporal information: two faces connected by the same track can be assumed to belong to the same person. Conversely, two faces detected at the same video frame at different locations do not belong to the same person. 
\item Geo-location: two faces captured from totally different geographic areas  are assumed to be from different people. 
\end{itemize}

\begin{table}
\centering \caption{Statistics of the Ego-Humans Dataset.} \label{table:egohumans}
\begin{tabular}{c|c}
\hline Collection period &  08/28 -- 10/26\\
\hline No. of days & 25 \\
\hline Video footage & \~40 hours\\
\hline Contextual info & GPS and Weather Data \\
\hline No. of face tracks & 15,000 \\
\hline No. of face images & 160,000 \\
\hline No. of generated face pairs & 4.9 million \\
\hline
\end{tabular}
\end{table}

Based on these constraints, we generate nearly 5 million face pairs, along with their {\em same/not same} labels. As detailed in the next section, preserving similarity of face pairs connected by the same track improves robustness to lighting and pose variation, and learning features to discriminate different individuals is important for the final facial attribute classification task. Table \ref{table:egohumans} summarizes the information about our data.  

\section{Facial Attribute Representation Learning}
In the previous section, we introduced our unique Ego-Humans dataset. Next, we describe how we use this data to build a rich visual representation for facial attribute classification, based on a deep network that encodes features related to facial similarity, as well as weather and location information, without requiring manual annotations in the pre-training stage.


\subsection{Learning Objective}
Our learning framework builds upon the millions of face pairs automatically generated based on face detection and landmark tracking as described in the previous section, along with weather and location information. Our goal is to learn good features for facial attribute classification by leveraging this data.
Specifically, given a face image $\bf{x}_i \in \mathcal{X}$  in the original pixel space, our goal is to obtain its associated facial representation $\bf{r}_i \in \mathcal{R}^N$, so that a facial attribute classifier can be constructed on top of $\bf{r}_i$ (e.g., via network fine-tuning with a small set of manual labels).


\begin{figure*}[t]
\centering 
\includegraphics[width=0.97\textwidth,height=0.37\textheight]{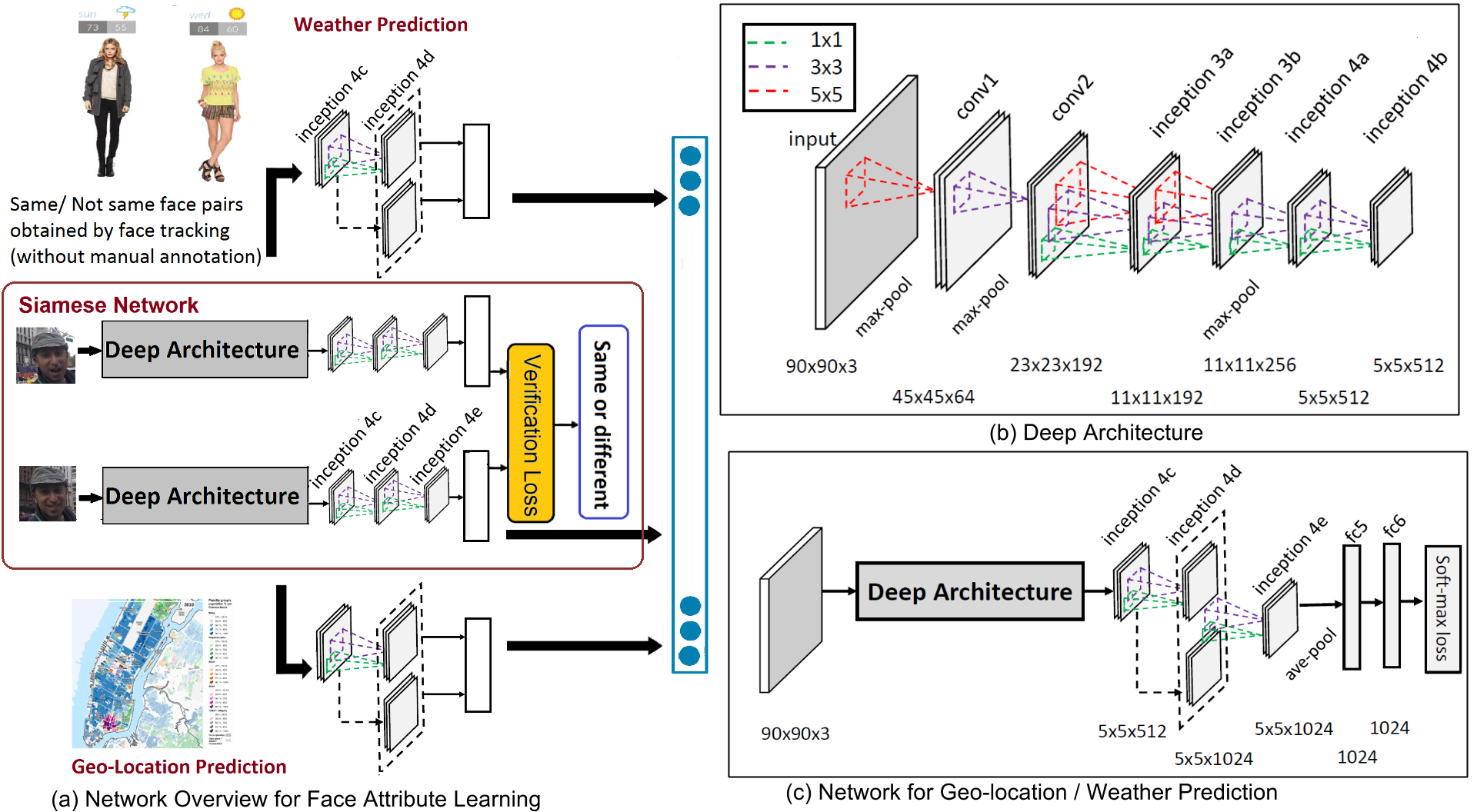}
\caption{(a) Overview of the proposed network for facial attribute learning. (b) The base deep architecture model. (c) The network for location/weather prediction }
\label{fig:overall}
\end{figure*}

In our learning algorithm, we have a training set $\mathcal{U}$ of $N_u$ face pairs,
$\mathcal{U} = \{(\mathbf{x}_i,\mathbf{x}_j);y_{i,j}\}$, where $y_{i,j} \in \{1,-1\}$ indicates  whether $(\bf{x}_i,\bf{x}_j)$ are images of the same person or not. In addition, we also have another two training sets $\mathcal{L}^{w}$ of $N_w$ images, $\mathcal{L}^{w} = \{\mathbf{x}_k; c_{k}\}^{N_w}_{k=1}$, where $c_{k} \in \{1,...,C_{w}\}$ indicates the label of weather; and $\mathcal{L}^{g} = \{\mathbf{x}_l; c_{l}\}^{N_g}_{g=1}$, where $c_{l} \in \{1,...,C_{g}\}$ indicates the label of geo-location. We use discretized values for weather and location labels as described in the previous section. 

The learned feature $\bf{r}_i$ should capture identity-related attributes (embedding in $\mathcal{U}$) and also preserve the high-level factors in $\mathcal{L}^{w}$ and $\mathcal{L}^{g}$. 
Towards this goal, the deep network is initially trained over $\mathcal{U}$ by minimizing the verification loss $\rm{d}_{e}(\mathord{\cdot})$ (to be described next) for face verification using a Siamese network structure. To learn high-level features from $\mathcal{L}^{w}$ and $\mathcal{L}^{g}$, we train weather and location networks independently by minimizing their own softmax loss functions. The two contextual networks are initialized by the weights from the verification-trained model on the bottom layers and fine-tuned with individual contextual labels. The feature $\bf{r}_i$ is the concatenation of the learned feature vectors of the top layer from each network and is further applied to train the facial attribute model.

\subsection{Deep Network Structure} To learn the embedding in $\mathcal{U}$ we design a Siamese network. A Siamese network consists of two {\em base networks} which share the same parameters. The Siamese structure is depicted in Figure \ref{fig:overall}(a). 
For our experiments, we take images with a size of $ 90\times90\times3$ as input.  The size of the face is constrained by the image quality and the resolution from the videos. 
The base network uses the GoogLeNet style architecture in~\cite{schroff2015facenet}.  
This deep architecture contains two convolutional layers and six layers of inception modules~\cite{szegedy2015going} as shown in Figure ~\ref{fig:overall}(b).  Due to the small input size, our architecture removes $5\times5$ filters from the inception models from layer inception 4a to inception 4d. The network contains around 4 million parameters.  
In the Siamese network, we connect the deep architecture with three inception modules and one fully connected layer. 

The weather and location models share the same base architecture as the Siamese network, but do not share parameters at the top layers. In particular, as illustrated in Figure \ref{fig:overall}(c), we feed the fully-connected layer with inception modules 4c and 4d. This allows us to capture more localized features in the weather and location models, while encoding more global similarity in the identity verification model.
In the three models (identity verification, weather, and location), the output feature vectors of the top fully connected layer are all 1024-dimensional vectors and are further concatenated to form the final facial attribute feature representation.
We implemented the network using the Caffe deep learning toolbox \cite{jia2014caffe}.
The complete network structure is shown in Figure \ref{fig:overall}(a).


\textbf{Loss Function}:
The Siamese network used to generate identity-related attribute features uses contrastive loss to preserve visual similarity of faces connected by the same track and dissimilarity to other tracks.
The contrastive loss $\rm{d}_{e}(\mathord{\cdot})$ is defined as:
\begin{eqnarray} \label{eq:loss:siamese}
\rm{d}_{e}(\mathbf{x}_i,\mathbf{x}_j,y_{i,j})=\mathbbm{1} (y_{i,j}=1) \rm{d}(\bf{x}_i,\bf{x}_j)+ \\ \nonumber \mathbbm{1} (y_{i,j}=-1)\rm{max}(\delta-d(\mathbf{x}_i,\mathbf{x}_j),0)
\end{eqnarray}
where $\mathbbm{1}(\mathord{\cdot})$ is the indicator function. This contrastive loss
penalizes the distance between $\bf{x}_i$ and $\bf{x}_j$ in “positive” mode, and pushes apart pairs in “negative” mode up to a minimum margin distance specified by the constant $\delta$. We use the $l_{2}$ norm for the distance measure.
The parameters of the network are updated using stochastic gradient descent (SGD) \cite{wilson2003general} by standard error back-propagation \cite{lecun1989backpropagation, rumelhart1988learning}. 
The weather and location prediction models use the softmax loss as mentioned earlier.

\textbf{Fine-Tuning for Attribute Learning.}
After we obtain our pre-trained model based on the optimization described previously, the next step is to use standard fine-tuning with images manually labeled with facial attribute labels. Additional output layers are added for fine-tuning and the cross-entropy loss is used for attribute classification. 
  

\section{Experiments}

\subsection{Ablation Studies on Wearable Data}
\begin{figure}[t]
\centering
\includegraphics[width=0.48\textwidth]{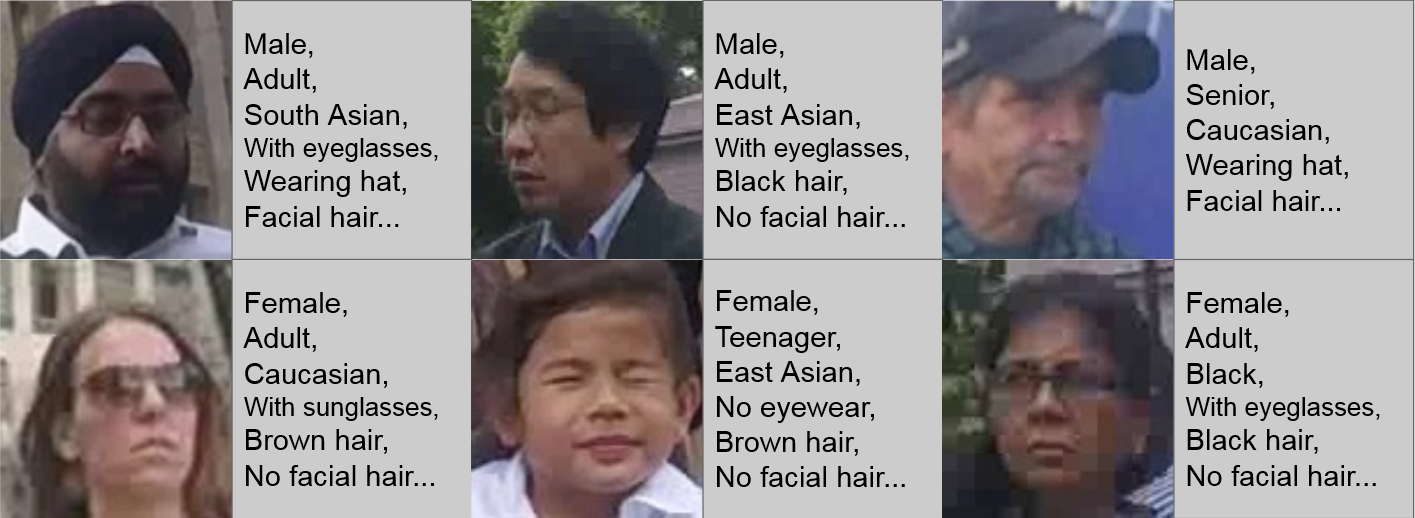}
\caption{Example of annotated wearable data with predicted attributes}
\label{fig:wearable_example}
\end{figure}

\begin{table}[t]
{\footnotesize
\begin{tabular}{l@{}|c|c|c|c|c|c|c|c|c}
\hline
\textbf{methods} &{\rotatebox[origin=c]{90}{Adult }} & {\rotatebox[origin=c]{90}{Bald }} & {\rotatebox[origin=c]{90}{Black }} & {\rotatebox[origin=c]{90}{Black Hair}} & {\rotatebox[origin=c]{90}{Brown Hair }} & {\rotatebox[origin=c]{90}{Caucasian }} & {\rotatebox[origin=c]{90}{East Asian }} & {\rotatebox[origin=c]{90}{Eyeglasses }} & {\rotatebox[origin=c]{90}{Facial Hair }} \\
\hline
random weights               & 84 & 90 & 86 & 71 & 71 & 81 & 84 & 81 & 83 \\
id(Ego-Humans)                             & 90 & 92 & 93 & 73 & 73 & 87 & 87 & 88 & 90 \\
\hline
\hline
& {\rotatebox[origin=c]{90}{Gender }} & {\rotatebox[origin=c]{90}{Hat }} & {\rotatebox[origin=c]{90}{No Eyewear }} & {\rotatebox[origin=c]{90}{Senior }}& {\rotatebox[origin=c]{90}{South Asian }} & {\rotatebox[origin=c]{90}{Sunglasses }} & {\rotatebox[origin=c]{90}{Teenager  }} & {\rotatebox[origin=c]{90}{Yellow hair }} & {\bf {\rotatebox[origin=c]{90}{Average}}}\\
\hline
random weights          & 82  & 82 & 86 & 85 & 83 & 89 & 85 & 85 & 82\\
id(Ego-Humans)            & 88  & 91 & 92 & 89 & 88 & 91 & 87 & 87 & {\bf 87}\\
\hline
\end{tabular}
}
\caption{Attribute prediction results of training the base net from scratch and with models after pre-training based on identity verification using the Ego-Humans dataset and the CASIA dataset. }
\label{tb:wearable_results}
\end{table}
In this section, we first analyze the effectiveness of each component of our network on our wearable dataset.
We have manually annotated 2714 images from 25 egocentric videos randomly selected from the data described in Section~\ref{sec:dataset}.   The faces in this dataset have large variations in pose and resolution.  Each annotated image contains seventeen facial attributes covering global attributes (e.g., gender, ethnicity, age) and local features (e.g., eyewear, hair color, hat).  All attributes are further categorized into binary class tasks.  For this dataset, we randomly select 80\% of the data as the training set and keep the rest for testing.  

{\bf Analysis of the Verification Model.}  
We first consider our base network (without the location and weather models). We evaluate the performance of training (fine-tuning) this network with the few available manual labeled examples, considering the following cases: 

1) training from scratch: The network is initialized with random weights and the global learning rate is set as 0.001.

2) id(Ego-Humans): Training with our pre-trained model based on identity verification with 5M image pairs automatically extracted from our Ego-Humans dataset. After pre-initializing the network with the weights learned from the verification models, we set the fine-tuning global learning rate with 0.0001, but with a learning rate in the top two layers of 10 times the global learning rate.

Both cases run through the whole wearable training data with 100 epochs in attribute learning. 
The results in Table~\ref{tb:wearable_results} demonstrate that the pre-trained model outperforms training from scratch by a large margin. This is not surprising, given the relatively small training size of the dataset and the large variations in pose and lighting. This demonstrates the richness of our verification model obtained from unlabeled egocentric videos. 

\begin{figure}[t]
\includegraphics[width=0.5\textwidth]{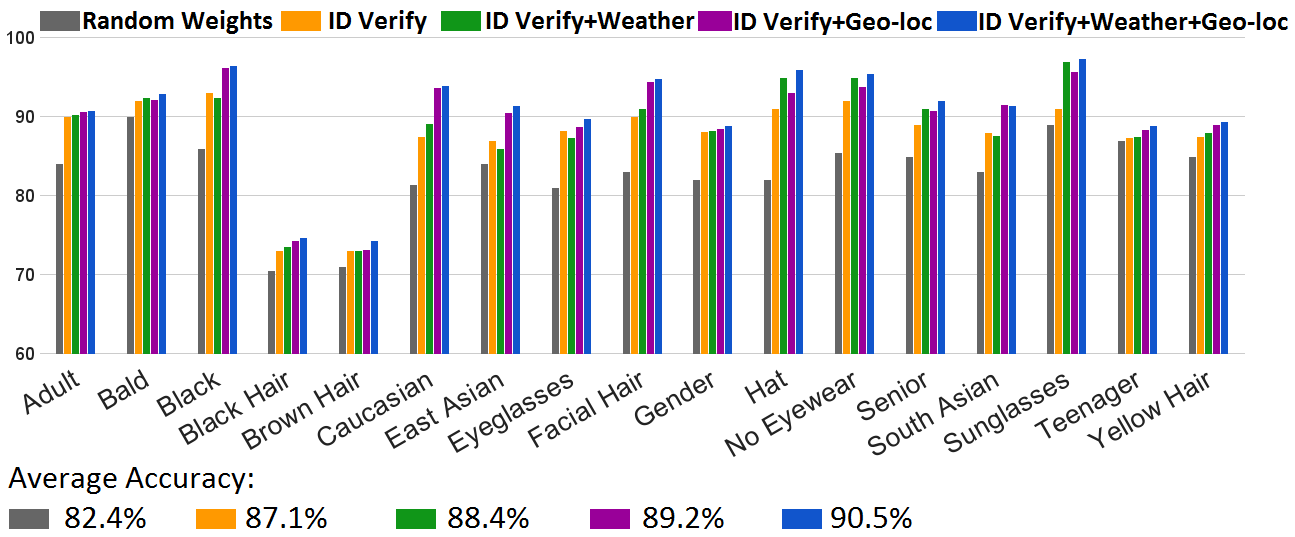}
\caption{Results of various baseline methods on the annotated wearable dataset. The embedding of location and weather net features help boost the performance, especially on ethnicity and non-identity related attributes. }
\label{fig:wearable_results}
\end{figure}

\begin{table*}[t]
{\footnotesize
\hfill{}
\begin{tabular}{c|l|c|c|c|c|c|c|c|c|c|c|c|c|c|c|c|c|c|c|c|c|c|}
\hline
\textbf{dataset} & \textbf{methods} & {\rotatebox[origin=c]{90}{5-o-shadow}} &  {\rotatebox[origin=c]{90}{Arch. Eyebrows}} & {\rotatebox[origin=c]{90}{Attractive}} & {\rotatebox[origin=c]{90}{Bags Un. Eyes}} & {\rotatebox[origin=c]{90}{Bald}} &  {\rotatebox[origin=c]{90}{Bangs}}  &  {\rotatebox[origin=c]{90}{Big Lips}} &   {\rotatebox[origin=c]{90}{Big Nose}}  &  {\rotatebox[origin=c]{90}{Black Hair}}  &   {\rotatebox[origin=c]{90}{Blond Hair}}  &   {\rotatebox[origin=c]{90}{Blurry}}  &   {\rotatebox[origin=c]{90}{Brown Hair}}  &   {\rotatebox[origin=c]{90}{Bussy Eyebrows}}  &   {\rotatebox[origin=c]{90}{Chubby}}  &   {\rotatebox[origin=c]{90}{Double Chin}}  &   {\rotatebox[origin=c]{90}{Eyeglasses}}  &   {\rotatebox[origin=c]{90}{Goatee}}   &   {\rotatebox[origin=c]{90}{Gray Hair}}  &   {\rotatebox[origin=c]{90}{Heavy Makeup}}  &   {\rotatebox[origin=c]{90}{H.Cheekbones}}  &   {\rotatebox[origin=c]{90}{Male}}\\
\hline
\multirow{5}{*}{LFWA} & FaceTracer & 70 & 67 & 71 & 65 & 77 & 72 & 68 & 73 & 76 & 88 & 73 & 62 & 67 & 67 & 70 & 90 & 69 & 78 & 88 & 77 & 84 \\
& PANDA-w & 64& 63& 70& 63& 82& 79& 64& 71& 78& 87& 70& 65& 63& 65& 64& 84& 65& 77& 86& 75& 86  \\
& PANDA-l   & 84& 79& 81& 80& 84& 84& 73& 79& 87& 94& 74& 74& 79& 69& 75& 89& 75& 81& 93& 86& 92 \\
& LNets+ANet & 84 & 82 & 83 & 83 & 88 & 88 & 75 & 81 & 90 & 97 & 74 & 77 & 82 & 73 & 78 & 95 & 78 & 84 & 95 & 88 & 94 \\
& Ours & 76 & 82 & 82 & 91 & 82 & 93 & 75 & 92 & 93 & 97 & 86 & 83 & 78 & 79 & 81 & 94 & 80 & 91 & 96 & 96 & 93 \\
\hline
\hline
\multirow{5}{*}{CelebA} & FaceTracer & 85 & 76 & 78 & 76 & 89 & 88 & 64 & 74 & 70 & 80 & 81 & 60 & 80 & 86 & 88 & 98 & 93 & 90 & 85 & 84 & 91 \\
& PANDA-w & 82 & 73 & 77 & 71 & 92 & 89 & 61 & 70 & 74 & 81 & 77 & 69 & 76 & 82 & 85 & 94 & 86 & 88 & 84 & 80 & 93 \\
& PANDA-l & 88 & 78 & 81 & 79 & 96 & 92 & 67 & 75 & 85 & 93 & 86 & 77 & 86 & 86 & 88 & 98 & 93 & 94 & 90 & 86 & 97 \\
& LNets+ANet & 91 & 79 & 81 & 79 & 98 & 95 & 68 & 78 & 88 & 95 & 84 & 80 & 90 & 91 & 92 & 99 & 95 & 97 & 90 & 87 & 98 \\
& Ours  & 84 & 87 & 84 & 87 & 92 & 96 & 78 & 91 & 84 & 92 & 91 & 81 & 93 & 89 & 93 & 97 & 92 & 95 & 96 & 95 & 96 \\
\hline
\hline
& & {\rotatebox[origin=c]{90}{Mouth S. O.}}  &    {\rotatebox[origin=c]{90}{Mustache}}  &   {\rotatebox[origin=c]{90}{Narrow Eyes}}  & {\rotatebox[origin=c]{90}{No Beard}}  &   {\rotatebox[origin=c]{90}{Oval Face}}  &    {\rotatebox[origin=c]{90}{Pale Skin}}  &   {\rotatebox[origin=c]{90}{Pointy Nose}}  &   {\rotatebox[origin=c]{90}{Reced. Hairline}}  &   {\rotatebox[origin=c]{90}{Rosy Cheeks}}  &   {\rotatebox[origin=c]{90}{Sideburns}}  &   {\rotatebox[origin=c]{90}{Smiling}}  &   {\rotatebox[origin=c]{90}{Straight Hair}}  &   {\rotatebox[origin=c]{90}{Wavy Hair}}  &  {\rotatebox[origin=c]{90}{Wear. Earings}}  &   {\rotatebox[origin=c]{90}{Wear. Hat}}  &   {\rotatebox[origin=c]{90}{Wear. Lipstick}}  &   {\rotatebox[origin=c]{90}{Wear. Necklace}}  &   {\rotatebox[origin=c]{90}{Wear. Necktie}}  &   {\rotatebox[origin=c]{90}{Young}}  & & {\rotatebox[origin=c]{90}{\textbf{Average}}}\\
\hline
\multirow{5}{*}{LFWA} & FaceTracer & 77 & 83 & 73 & 69 & 66 & 70 & 74 & 63 & 70 & 71 & 78 & 67 & 62 & 88 & 75 & 87 & 81 & 71 & 80 & & 74\\
& PANDA-w & 74 & 77 & 68 & 63 & 64 & 64 & 68 & 61 & 64 & 68 & 77 & 68 & 63 & 85 & 78 & 83 & 79 & 70 & 76  & & 71 \\
& PANDA-l & 78 & 87 & 73 & 75 & 72 & 84 & 76 & 84 & 73 & 76 & 89 & 73 & 75 & 92 & 82 & 93 & 86 & 79 & 82 & & 81 \\
& LNets+ANet & 82 & 92 & 81 & 79 & 74 & 84 & 80 & 85 & 78 & 77 & 91 & 76 & 76 & 94 & 88 & 95 & 88 & 79 & 86 & & 84 \\
& Ours & 94 & 83 & 79 & 75 & 84 & 87 & 93 & 86 & 81 & 77 & 97 & 76 & 89 & 96 & 86 & 97 & 95 & 80 & 89 & & \textbf{87}\\
\hline
\hline
\multirow{5}{*}{CelebA} & FaceTracer & 87 & 91 & 82 & 90 & 64 & 83 & 68 & 76 & 84 & 94 & 89 & 63 & 73 & 73 & 89 & 89 & 68 & 86 & 80 & & 81 \\
& PANDA-w & 82 & 83 & 79 & 87 & 62 & 84 & 65 & 82 & 81 & 90 & 89 & 67 & 76 & 72 & 91 & 88 & 67 & 88 & 77 & & 79 \\
& PANDA-l & 93 & 93 & 84 & 93 & 65 & 91 & 71 & 85 & 87 & 93 & 92 & 69 & 77 & 78 & 96 & 93 & 67 & 91 & 84 & & 85 \\
& LNets+ANet & 92 & 95 & 81 & 95 & 66 & 91 & 72 & 89 & 90 & 96 & 92 & 73 & 80 & 82 & 99 & 93 & 71 & 93 & 87 & & 87 \\
& Ours    & 97 & 90 & 79 & 90 & 79 & 85 & 77 & 84 & 96 & 92 & 98 & 75 & 85 & 91 & 96 & 92 & 77 & 84 & 86 & & \textbf{88} \\

\hline
\end{tabular}
}
\hfill{}
\caption{Performance comparison with state of the art methods on 40 binary facial attributes}
\label{tb:lfwa}
\end{table*}

{\bf Analysis of the Geo-Location and Weather Models.} 
Now we evaluate the benefit of features learned from geo-location and weather models.  We perform experiments on the annotated wearable dataset by fine-tuning the network, considering the identity verification model only (id-verify), the inclusion of geo-location (id-verify + geo-loc), weather (id-verify + weather) and geo-location and weather models concatenated (id-verify + weather + geo-loc). The results are shown in Figure~\ref{fig:wearable_results}.  
The performance indicates average improvements of 2\%, 2\% and 4\% when concatenating the base-net with the features fine-tuned from the geo-location, weather, and geo-location + weather, respectively.  The geo-location model provides more complementary information to the verification network on ethnicities like East Asian and South Asian. And the weather model adds in weights for non-identity-related but weather-related attributes like sunglasses and hat. Figure \ref{fig:wearable_example} illustrates some examples of attribute prediction in our data.

\subsection{Comparison with the State of the Art}
In this section we evaluate the effectiveness of our network with quantitative results on two standard facial attribute datasets, CelebA and LFWA, constructed  based on face datasets CelebFaces ~\cite{sun2014deep} and LFW ~\cite{huang2007labeled}, respectively.  Both datasets have forty binary facial attributes, as listed in Table ~\ref{tb:lfwa}. We use the exact same partition of data as in ~\cite{liu2015faceattributes}: 160k images of CelebA are used to fine-tune the network. In the remaining 40k CelebA images and the LFWA dataset, 50\% of the images are used to extract the learned top-layer fc features from the network and to train a linear SVM classifier for attribute classification, and the other 50\% are used for testing.

We evaluate the performance of our network on the two datasets with four state-of-the-art methods: FaceTracer~\cite{kumar2008facetracer}, two versions of PANDA~\cite{zhang2014panda} network, PANDA-w and PANDA-l, based on the setting described in~\cite{liu2015faceattributes}; and LNet+ANet~\cite{liu2015faceattributes}. The same data was used for all approaches.  FaceTracer utilizes hand-crafted features (HOG + color histogram) on face functional regions to train an SVM classifier.
LNet+ANet uses a massive set of images with manually labeled identities for pre-training and cascades two networks to automatically detect the face region and consequently learn the facial attributes from the detected part.Apart from LNet+ANet, all the methods obtain cropped faces externally either from given landmark points (FaceTracer and PANDA-l) or based on off-the-shelf detection (PANDA-w and ours).  


As shown in Table \ref{tb:lfwa}, our approach significantly outperforms the four other methods on LFWA and reaches comparable performance with LNet+ANet on CelebA on average score, without using manual labeling in the pre-training stage. Our approach achieves better results than the prior methods on most of the forty attributes. 
\subsection{Visual Attribute Discovery}


The quantitative results for the above three datasets show that the pre-trained models on identity verification, geo-location, and weather classification boost the prediction of facial attributes despite the fact they are not explicitly trained for attribute classification.  To better understand the attribute-related contextual information the pre-trained nets have learned, we show some qualitative examples of the top activated neurons in the pre-training phase.
\begin{figure}[t]
\centering
{\includegraphics[width=0.49\textwidth]{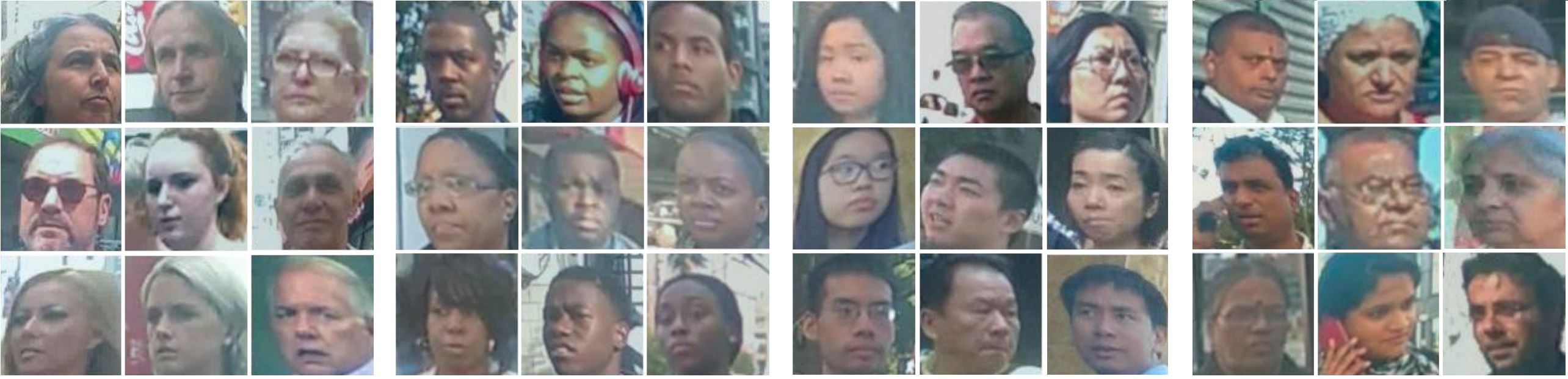}}
\caption{Top selected images in each class with max activations on layer inception 4e after pre-training loc-net. The discretized geo-location classes according to census from left to right are: White, Black, Asian, and Indian.}
\label{fig:vis1}
\end{figure}
\begin{figure}[t]
\centering
{\includegraphics[width=0.49\textwidth]{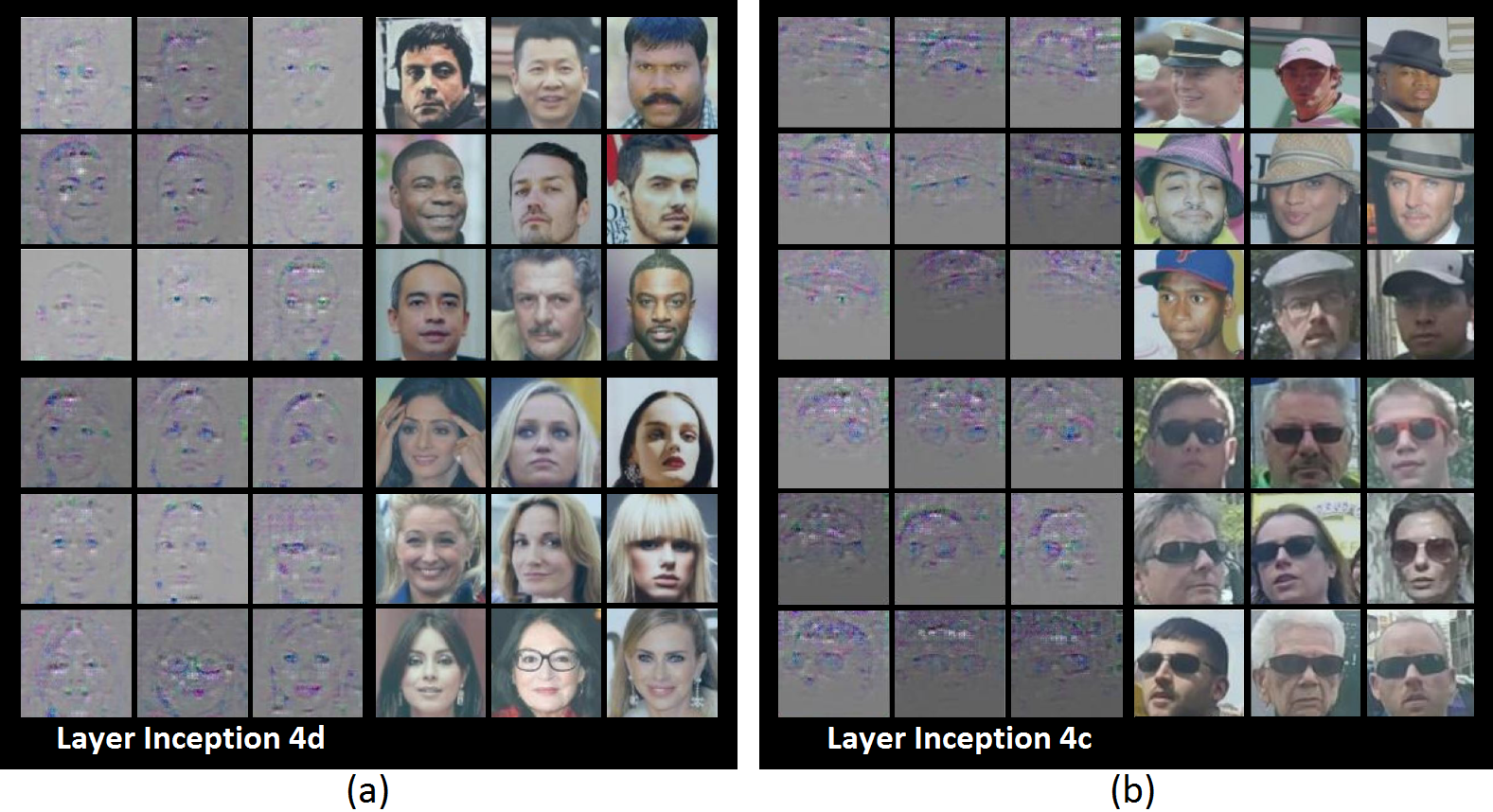}}
\caption{Visualized feature of top ranked neurons in models after pre-training identity verification (a) and pre-training weather classification (b). Best viewed in electronic form.}
\label{fig:vis2}
\end{figure}

Given a layer in the pre-trained net, for example inception 4d in loc-net,  we get top class-related neurons with high activations across all images within a class.  For each selected neuron, we further select the most related image with highest value across the whole dataset.  Figure~\ref{fig:vis1} shows the corresponding selected images of the top nine neurons in each class. We can see that the selected images in different geo-locations are from different races, which means the neurons in loc-net are learning strong priors about the concept of ethnicity in the location classification. 

To better visualize attributes discovered by neurons,  we construct the deconvnet framework following the ideas in~\cite{zeiler2014visualizing} and project selected top neurons back to the input pixel space.  
Figure~\ref{fig:vis2} presents the visualized features of the top nine neurons of specific layers after pre-training identity verification and weather classification separately.  Recovering the whole face contour with clear discriminative parts such as the eyes and mouth, the visualized results of the selected neurons in the verification model from layer inception 4d in Fig~\ref{fig:vis2}(a) reveal that the neurons capture global identity-related face features.  Therefore, facial attributes that are intrinsic to the identity, such as ``gender", can be discovered by the network. 
The illustrated neurons in the weather model are from layer inception 4c and 4d.  The visualizations of the selected neurons partially recover the upper face and focus on similar local components.  By capturing local attributes such as ``sunglasses" or ``hat", the visualization explicitly demonstrates that the pre-trained weather model provides complementary features on identity-non-related attributes to the model from identity verification.

\section{Discussion}
In the next few years, the amount of egocentric video associated with contextual information will grow significantly. As an example, many police officers around the world are already using body-worn cameras in patrol operations. 
This growth may be even greater as wearable devices become mainstream among ordinary people.
We believe our work offers novel ways to learn rich facial representations from the ever-growing pool of unlabeled egocentric videos with contextual data. Although we have considered only the case of people walking across different neighborhoods of a city, our method could be applied at different geo-scales (e.g. worldwide) to capture larger variations. 

One could argue that the face pairs generated by our approach inherit some bias. To the contrary, we have shown that in practice this is not an issue. In fact, we observed that faces across the same track exhibit large variations in pose and lighting, helping our approach to be more robust against these factors. 

We would like to point out that our approach only requires location and weather data at the training stage. Although this contextual information could be useful at test time to improve accuracy, it may not always be available. 

Finally, we are currently applying our approach to learn representations for fine-grained clothing attribute classification \cite{chen2015deep}, as weather and location clearly influence clothing choices. 
By learning with diverse contextual information, the framework could be also applied to other high-level analysis tasks such as urban perception \cite{urbanECCV2014}.

\vspace{0.1in}

\section{Conclusions}
In this paper we have proposed a novel deep learning framework for learning facial attributes. Different from previous approaches, our method can capture good representations/features for facial  attributes by exploiting videos and contextual data (geo-location and weather) captured by a wearable sensor as the person walks. The proposed framework can leverage 
the rich appearance of faces from tens of thousands of casual walkers at different locations and lighting conditions without requiring the cost of manual labels. We demonstrate
our approach in several real-world datasets, showing substantial improvement over
other baselines.

{\small
\bibliographystyle{ieee}
\bibliography{egbib}

\begin{thebibliography}{10}\itemsep=-1pt

\bibitem{LSM2015}
P.~Agrawal, J.~Carreira, and J.~Malik.
\newblock Learning to see by moving.
\newblock In {\em Proc. ICCV}, 2015.

\bibitem{berg-poof-cvpr2013}
T.~Berg and P.~N. Belhumeur.
\newblock {POOF}: {P}art-{B}ased {O}ne-vs-{O}ne {F}eatures for fine-grained
  categorization, face verification, and attribute estimation.
\newblock In {\em Proc. CVPR}, 2013.

\bibitem{DBLP:journals/tcsv/BetancourtMRR15}
A.~Betancourt, P.~Morerio, C.~S. Regazzoni, and M.~Rauterberg.
\newblock The evolution of first person vision methods: {A} survey.
\newblock {\em {IEEE} Trans. Circuits Syst. Video Techn.}, 25(5):744--760,
  2015.

\bibitem{hchen_gallagher_girod_cvpr_13_firstnames}
H.~Chen, A.~Gallagher, and B.~Girod.
\newblock What's in a name: First names as facial attributes.
\newblock In {\em Proc. CVPR}, 2013.

\bibitem{chen2015deep}
Q.~Chen, J.~Huang, R.~Feris, L.~M. Brown, J.~Dong, and S.~Yan.
\newblock Deep domain adaptation for describing people based on fine-grained
  clothing attributes.
\newblock In {\em Proc. CVPR}, 2015.

\bibitem{imagenet_cvpr09}
J.~Deng, W.~Dong, R.~Socher, L.-J. Li, K.~Li, and L.~Fei-Fei.
\newblock {ImageNet: A Large-Scale Hierarchical Image Database}.
\newblock In {\em Proc. CVPR}, 2009.

\bibitem{doersch2015unsupervised}
C.~Doersch, A.~Gupta, and A.~A. Efros.
\newblock Unsupervised visual representation learning by context prediction.
\newblock In {\em Proc. ICCV}, 2015.

\bibitem{Doersch:2012:MPL:2185520.2185597}
C.~Doersch, S.~Singh, A.~Gupta, J.~Sivic, and A.~A. Efros.
\newblock What makes paris look like paris?
\newblock {\em ACM Trans. Graph.}, 31(4):101:1--101:9, July 2012.

\bibitem{DBLP:conf/cvpr/FathiHR12}
A.~Fathi, J.~K. Hodgins, and J.~M. Rehg.
\newblock Social interactions: {A} first-person perspective.
\newblock In {\em Proc. CVPR}, 2012.

\bibitem{Fathi:2012:LRD:2402940.2402964}
A.~Fathi, Y.~Li, and J.~M. Rehg.
\newblock Learning to recognize daily actions using gaze.
\newblock In {\em Proc. ECCV}, 2012.

\bibitem{Feris:2014:APS:2578726.2578732}
R.~Feris, R.~Bobbitt, L.~Brown, and S.~Pankanti.
\newblock Attribute-based people search: Lessons learnt from a practical
  surveillance system.
\newblock In {\em Proc. ICMR}, 2014.

\bibitem{gronat2013learning}
P.~Gronat, G.~Obozinski, J.~Sivic, and T.~Pajdla.
\newblock Learning and calibrating per-location classifiers for visual place
  recognition.
\newblock In {\em Proc. CVPR}, 2013.

\bibitem{hays2008im2gps}
J.~Hays, A.~Efros, et~al.
\newblock Im2gps: estimating geographic information from a single image.
\newblock In {\em Proc. CVPR}, 2008.

\bibitem{Hinton:2006:FLA:1161603.1161605}
G.~E. Hinton, S.~Osindero, and Y.-W. Teh.
\newblock A fast learning algorithm for deep belief nets.
\newblock {\em Neural Comput.}, 18(7):1527--1554, July 2006.

\bibitem{huang2007labeled}
G.~B. Huang, M.~Ramesh, T.~Berg, and E.~Learned-Miller.
\newblock Labeled faces in the wild: A database for studying face recognition
  in unconstrained environments.
\newblock Technical report, Technical Report 07-49, University of
  Massachusetts, Amherst, 2007.

\bibitem{geofacial}
M.~T. Islam, C.~Greenwell, R.~Souvenir, and N.~Jacobs.
\newblock {Large-Scale Geo-Facial Image Analysis}.
\newblock {\em {EURASIP Journal on Image and Video Processing (JIVP)}},
  2015(1):1--14, June 2015.

\bibitem{islam2015face2gps}
M.~T. Islam, S.~Workman, and N.~Jacobs.
\newblock {Face2GPS: Estimating Geographic Location from Facial Features}.
\newblock In {\em Proc. ICIP}, 2015.

\bibitem{DBLP:journals/corr/JayaramanG15}
D.~Jayaraman and K.~Grauman.
\newblock Learning image representations equivariant to ego-motion.
\newblock In {\em Proc. ICCV}, 2015.

\bibitem{jia2014caffe}
Y.~Jia, E.~Shelhamer, J.~Donahue, S.~Karayev, J.~Long, R.~Girshick,
  S.~Guadarrama, and T.~Darrell.
\newblock Caffe: Convolutional architecture for fast feature embedding.
\newblock In {\em Proc. ACM Multimedia}, 2014.

\bibitem{kiapourbuy}
M.~H. Kiapour, X.~Han, S.~Lazebnik, A.~C. Berg, and T.~L. Berg.
\newblock Where to buy it: Matching street clothing photos in online shops.
\newblock In {\em Proc. ICCV}, 2015.

\bibitem{krizhevsky2012imagenet}
A.~Krizhevsky, I.~Sutskever, and G.~E. Hinton.
\newblock Imagenet classification with deep convolutional neural networks.
\newblock In {\em NIPS}, 2012.

\bibitem{kumar2008facetracer}
N.~Kumar, P.~Belhumeur, and S.~Nayar.
\newblock Facetracer: A search engine for large collections of images with
  faces.
\newblock In {\em Proc. ECCV}, 2008.

\bibitem{kumar2011describable}
N.~Kumar, A.~C. Berg, P.~N. Belhumeur, and S.~K. Nayar.
\newblock Describable visual attributes for face verification and image search.
\newblock {\em Pattern Analysis and Machine Intelligence, IEEE Transactions
  on}, 33(10):1962--1977, 2011.

\bibitem{layne2014re}
R.~Layne, T.~M. Hospedales, and S.~Gong.
\newblock Re-id: Hunting attributes in the wild.
\newblock In {\em BMVC}, 2014.

\bibitem{DBLP:conf/icml/LeRMDCCDN12}
Q.~V. Le, M.~Ranzato, R.~Monga, M.~Devin, G.~Corrado, K.~Chen, J.~Dean, and
  A.~Y. Ng.
\newblock Building high-level features using large scale unsupervised learning.
\newblock In {\em Proc. ICML}, 2012.

\bibitem{lecun1989backpropagation}
Y.~LeCun, B.~Boser, J.~S. Denker, D.~Henderson, R.~E. Howard, W.~Hubbard, and
  L.~D. Jackel.
\newblock Backpropagation applied to handwritten zip code recognition.
\newblock {\em Neural computation}, 1(4):541--551, 1989.

\bibitem{lee2015predicting}
S.~Lee, H.~Zhang, and D.~J. Crandall.
\newblock Predicting geo-informative attributes in large-scale image
  collections using convolutional neural networks.
\newblock In {\em WACV}, 2015.

\bibitem{DBLP:journals/ijcv/LeeG15}
Y.~J. Lee and K.~Grauman.
\newblock Predicting important objects for egocentric video summarization.
\newblock {\em International Journal of Computer Vision}, 114(1):38--55, 2015.

\bibitem{twobirdsonestone}
Y.~Li, R.~Wang, H.~Liu, H.~Jiang, S.~Shan, and X.~Chen.
\newblock Two birds, one stone: Jointly learning binary code for large-scale
  face image retrieval and attributes prediction.
\newblock In {\em Proc. CVPR}, 2015.

\bibitem{Li_2015_CVPR}
Y.~Li, Z.~Ye, and J.~M. Rehg.
\newblock Delving into egocentric actions.
\newblock In {\em Proc. CVPR}, 2015.

\bibitem{lin2013cross}
T.-Y. Lin, S.~Belongie, and J.~Hays.
\newblock Cross-view image geolocalization.
\newblock In {\em Proc. CVPR}, 2013.

\bibitem{liu2012street}
S.~Liu, Z.~Song, G.~Liu, C.~Xu, H.~Lu, and S.~Yan.
\newblock Street-to-shop: Cross-scenario clothing retrieval via parts alignment
  and auxiliary set.
\newblock In {\em Proc. CVPR}, 2012.

\bibitem{liu2015faceattributes}
Z.~Liu, P.~Luo, X.~Wang, and X.~Tang.
\newblock Deep learning face attributes in the wild.
\newblock In {\em Proc. ICCV}, 2015.

\bibitem{Luo:2013:DSA:2586117.2587104}
P.~Luo, X.~Wang, and X.~Tang.
\newblock A deep sum-product architecture for robust facial attributes
  analysis.
\newblock In {\em Proc. ICCV}, 2013.

\bibitem{Mobahi:2009:DLT:1553374.1553469}
H.~Mobahi, R.~Collobert, and J.~Weston.
\newblock Deep learning from temporal coherence in video.
\newblock In {\em Proc. ICML}, 2009.

\bibitem{urbanECCV2014}
V.~Ordonez and T.~L. Berg.
\newblock Learning high-level judgments of urban perception.
\newblock In {\em Proc. ECCV}, 2014.

\bibitem{Rogez_2015_CVPR}
G.~Rogez, J.~S. Supancic, III, and D.~Ramanan.
\newblock First-person pose recognition using egocentric workspaces.
\newblock In {\em Proc. CVPR}, 2015.

\bibitem{rumelhart1988learning}
D.~E. Rumelhart, G.~E. Hinton, and R.~J. Williams.
\newblock Learning representations by back-propagating errors.
\newblock {\em Cognitive Modeling}, 5(3):1, 1988.

\bibitem{schroff2015facenet}
F.~Schroff, D.~Kalenichenko, and J.~Philbin.
\newblock Facenet: A unified embedding for face recognition and clustering.
\newblock In {\em Proc. CVPR}, 2015.

\bibitem{DBLP:conf/cvpr/ShiHX15}
Z.~Shi, T.~M. Hospedales, and T.~Xiang.
\newblock Transferring a semantic representation for person re-identification
  and search.
\newblock In {\em Proc. CVPR}, 2015.

\bibitem{Siddiquie:2011:IRR:2191740.2192123}
B.~Siddiquie, R.~S. Feris, and L.~S. Davis.
\newblock Image ranking and retrieval based on multi-attribute queries.
\newblock In {\em Proc. CVPR}, 2011.

\bibitem{DBLP:conf/icml/SrivastavaMS15}
N.~Srivastava, E.~Mansimov, and R.~Salakhutdinov.
\newblock Unsupervised learning of video representations using lstms.
\newblock In {\em Proc. ICML}, 2015.

\bibitem{sun2014deep}
Y.~Sun, Y.~Chen, X.~Wang, and X.~Tang.
\newblock Deep learning face representation by joint
  identification-verification.
\newblock In {\em NIPS}, 2014.

\bibitem{DBLP:conf/cvpr/SunWT15}
Y.~Sun, X.~Wang, and X.~Tang.
\newblock Deeply learned face representations are sparse, selective, and
  robust.
\newblock In {\em Proc. CVPR}, 2015.

\bibitem{szegedy2015going}
C.~Szegedy, W.~Liu, Y.~Jia, P.~Sermanet, S.~Reed, D.~Anguelov, D.~Erhan,
  V.~Vanhoucke, and A.~Rabinovich.
\newblock Going deeper with convolutions.
\newblock In {\em Proc. CVPR}, 2015.

\bibitem{TangPFFB15}
K.~Tang, P.~Manohar, F.~Li, F.~Rob, and D.~B. Lubomir.
\newblock Improving image classification with location context.
\newblock In {\em Proc. ICCV}, 2015.

\bibitem{rita2015}
P.~Varini, G.~Serra, and R.~Cucchiara.
\newblock Egocentric video summarization of cultural tour based on user
  preferences.
\newblock In {\em Proc. ACM Multimedia}, 2015.

\bibitem{Vincent:2008:ECR:1390156.1390294}
P.~Vincent, H.~Larochelle, Y.~Bengio, and P.-A. Manzagol.
\newblock Extracting and composing robust features with denoising autoencoders.
\newblock In {\em Proc. ICML}, 2008.

\bibitem{DBLP:journals/corr/WangG15a}
X.~Wang and A.~Gupta.
\newblock Unsupervised learning of visual representations using videos.
\newblock In {\em Proc. ICCV}, 2015.

\bibitem{wilson2003general}
D.~R. Wilson and T.~R. Martinez.
\newblock The general inefficiency of batch training for gradient descent
  learning.
\newblock {\em Neural Networks}, 16(10):1429--1451, 2003.

\bibitem{Wiskott:2002:SFA:638940.638941}
L.~Wiskott and T.~J. Sejnowski.
\newblock Slow feature analysis: Unsupervised learning of invariances.
\newblock {\em Neural Comput.}, 14(4):715--770, Apr. 2002.

\bibitem{xiong2013supervised}
X.~Xiong and F.~De~la Torre.
\newblock Supervised descent method and its applications to face alignment.
\newblock In {\em Proc. CVPR}, 2013.

\bibitem{DBLP:conf/cvpr/ChengFPC14}
S.~P. Yu~Cheng, Quanfu~Fan and A.~N. Choudhary.
\newblock Temporal sequence modeling for video event detection.
\newblock In {\em Proc. CVPR}, 2014.

\bibitem{zeiler2014visualizing}
M.~D. Zeiler and R.~Fergus.
\newblock Visualizing and understanding convolutional networks.
\newblock In {\em Proc. ECCV}, 2014.

\bibitem{Zeiler10deconvolutionalnetworks}
M.~D. Zeiler, D.~Krishnan, G.~W. Taylor, and R.~Fergus.
\newblock Deconvolutional networks.
\newblock In {\em Proc. CVPR}, 2010.

\bibitem{zhang2014panda}
N.~Zhang, M.~Paluri, M.~Ranzato, T.~Darrell, and L.~Bourdev.
\newblock Panda: Pose aligned networks for deep attribute modeling.
\newblock In {\em Proc. CVPR}, 2014.

\bibitem{socialrelation_ICCV2015}
Z.~Zhang, P.~Luo, C.-C. Loy, and X.~Tang.
\newblock Learning social relation traits from face images.
\newblock In {\em Proc. ICCV}, 2015.

\bibitem{DBLP:journals/corr/ZhaoMGL15}
J.~Zhao, M.~Mathieu, R.~Goroshin, and Y.~LeCun.
\newblock Stacked what-where auto-encoders.
\newblock {\em CoRR}, abs/1506.02351, 2015.

\bibitem{LNCS86910519}
B.~Zhou, L.~Liu, A.~Oliva, and A.~Torralba.
\newblock Recognizing city identity via attribute analysis of geo-tagged
  images.
\newblock In {\em Proc. ECCV}, 2014.

\end{thebibliography}
}

\end{document}